\documentclass[11pt,a4paper]{article}
\usepackage[hyperref]{eacl2021}
\usepackage{times}
\usepackage{latexsym}

\usepackage{tabularx}

\usepackage{url}
\usepackage{wrapfig}
\usepackage{graphicx}
\usepackage{enumitem}
\usepackage{amsfonts}
\usepackage{amsmath}
\usepackage{multirow}
\usepackage{subcaption}
\usepackage{booktabs}
\usepackage[T1]{fontenc}
\usepackage{color}
\usepackage{multirow}
\usepackage{wrapfig}
\usepackage{graphicx}
\usepackage{blindtext}

\aclfinalcopy 

\newcounter{tbsnr}
\newenvironment{tbs}
{\addtocounter{tbsnr}{1}\par\bigskip\noindent\fbox{\thetbsnr}
	\hspace*{\fill}\begin{minipage}{7cm}\tt}
	{\end{minipage}\hspace*{\fill}\bigskip}
\newcommand{\tb}[1]{\begin{tbs}{#1}\end{tbs}}

\newcommand{\cut}[1]{}

\newcommand{\rb}[1]{\textcolor{blue}{\textbf{RB: #1}}}
\newcommand{\at}[1]{\textcolor{magenta}{\textbf{#1}}}

\title{The Interplay of Task Success and Dialogue Quality:\\ An in-depth Evaluation in Task-Oriented Visual Dialogues}

\cut{
\author{
	Alberto Testoni $^{1}$ , Raffaella Bernardi $^{1,2}$\\
	$^{1}$Department of Information Engineering and Computer Science - DISI - University of Trento \\
	$^{2}$Center for Mind/Brain Sciences - CIMeC - University of Trento\\
	\texttt{\{alberto.testoni|raffaella.bernardi\}@unitn.it}\\
	
}
}

\author{Alberto Testoni \\
  DISI - University of Trento \\
  Trento - Italy \\
  \texttt{alberto.testoni@unitn.it} \\\And
  Raffaella Bernardi \\
  CIMeC and DISI - University of Trento\\
  Rovereto (TN) - Italy \\
  \texttt{raffaella.bernardi@unitn.it} \\}

\date{}

\begin{document}
\maketitle
\begin{abstract}
  When training a model on referential dialogue guessing games, the best model is
  usually chosen based on its task success.\@ We show that in the
  popular end-to-end approach, this choice prevents the model from
  learning to generate linguistically richer dialogues, since the
  acquisition of language proficiency takes longer than learning the
  guessing task.\@ By comparing models playing different  games (GuessWhat, GuessWhich, and Mutual Friends), we show
  that this discrepancy is model- and task-agnostic.\@ We
  investigate whether and when better language quality could lead to higher task success. We show that in GuessWhat, models could
  increase their accuracy if they learn to ground, encode, and decode also
  words that do not occur frequently in the training set.
\end{abstract}

\section{Introduction}
\label{sec:intro}


A good dialogue model should generate utterances that are
indistinguishable from human dialogues
\citep{liu-etal-2016-evaluate,li-etal-2017-adversarial}. This holds
for both chit-chat, open-domain, and task-oriented dialogues.  While
chit-chat dialogue systems are usually evaluated by analysing the
quality of their
dialogues~\cite{lowe-etal-2017-towards,see-etal-2019-makes},
task-oriented dialogue models are evaluated on their task success
and it is common practice to choose the best model based only on the
task success metric. We explore whether this choice prevents the
system from learning better linguistic skills.

Important progress has been made on the development of such
conversational agents.  The boost is mostly due to the introduction
of the encoder-decoder framework \cite{suts:seq14} which allows
learning directly from raw data to both understand and generate
utterances. The framework has been found to be promising both for
chit-chat \cite{Vinyals2015ANC} and task-oriented dialogues
\cite{deal:lewi17}, and it has been further extended to develop agents
that can communicate through natural language about visual content
\cite{imag:mosta17,visdial,guesswhat_game}.  Several dialogue tasks
have been proposed as \emph{referential guessing games} in which an
agent (the Q-bot) asks questions to another agent (the A-bot) and has
to guess the referent (e.g., a specific object depicted in the image)
they have been speaking about
\cite{guesswhat_game,visdial_rl,he:lear17,haber-etal-2019-photobook,ilinykh-etal-2019-tell,udag19}. We
are interested in understanding the interplay between the learning
processes behind these two sub-tasks: generating questions and
guessing the referent.

Shekhar et al.\@ \shortcite{shekhar-etal-2019-beyond} have compared
models on GuessWhat and have shown that task success (TS) does not
correlate with the quality of machine-generated dialogues. First of all, we check whether
this result is task-agnostic by carrying out a comparative analysis of
models playing different referential games.  We choose a task in which
visual grounding happens during question generation (GuessWhat, \citealt{guesswhat_game}); a task in which it happens only in the
guessing phase (GuessWhich, \citealt{visdial_rl}), and a task that
is only based on language (MutualFriends, \citealt{he:lear17}).  We
introduce a linguistic metric, Linguistic Divergence (LD), that, by
assembling various metrics used in the
literature~\citep{shekhar-etal-2019-beyond,mura:impr19,van-miltenburg-etal-2018-measuring},
measures how much the language generated by computational models
differs, on the surface level, from the one
used by humans. We consider LD to be a proxy of the quality of machine-generated dialogues. 

\cut{More specifically, we have taken the most commonly known deficiencies of generative models as our starting point to evaluate dialogue quality, namely questions/token repetitions,  poor vocabulary, inconsistency in word usage. Hence, we have focused on comparing models against existing metrics that capture these characteristics of dialogues and propose a new metric (LD) that aggregates them; based on this metric we consider a dialogue of good quality if such deficiencies are less present.} 

For each task, we compare State-Of-The-Art (SOTA) models
against their TS and LD.  In the core part of the paper, we study the
relationship between the learning process behind TS and LD by comparing
model performance across epochs and by downsizing the training
set. Finally, we study whether and when a lower LD (i.e., the generated dialogues are more similar to humans) could help
reach a higher TS.

Our results confirm that models performing similarly on TS differ quite a lot on their conversational skills, as claimed in Shekhar et al.  \shortcite{shekhar-etal-2019-beyond}
for models evaluated on the GuessWhat game.  Furthermore, we show that:



\begin{itemize}
\item SOTA models are much faster in achieving high performance on
  the guessing task compared to reaching a high dialogue quality (i.e., low LD).
Hence, choosing the best
  model on task success prevents the model from reaching better
  conversational skills; 
\item SOTA models mostly use very frequent words; this limited vocabulary
  is sufficient for succeeding in a high number of games;
\item in GusseWhat, a higher TS could be reached if the model learns to
  use also less frequent words.


\end{itemize}


\cut{
\begin{figure}[t]
	\begin{center}
		\begin{tabular}{c|c|c}
			{\textbf{GuessWhat?!}} &
			{\textbf{GuessWhich}} &
			{\textbf{Mutual Friends}} 
			\\\hline
			\begin{tabular}{c}
				Is it a person? No\\
				Is it a dog? No\\
				Is it a cat? Yes\\
				The one on the left? Yes
			\end{tabular}
			&
			\begin{tabular}{c}
				How many people are? Two\\
				How old are they? They are young\\
				Is it indoor? Yes\\
				Is there a TV? No
			\end{tabular}
			& 
			\multirow{2}{*}{
				\begin{tabular}{p{3.5cm}}
					A:Hi! My friends like to do stuff outdoors. \\
					B:Not a single one of mine do! \\
					A:ok so lets go with the indoor friends
					B:Many of my friends work at Google \\
				\end{tabular}
			}
			\\\cline{1-2}
			\begin{tabular}{c}
				\\\includegraphics[scale=0.13]{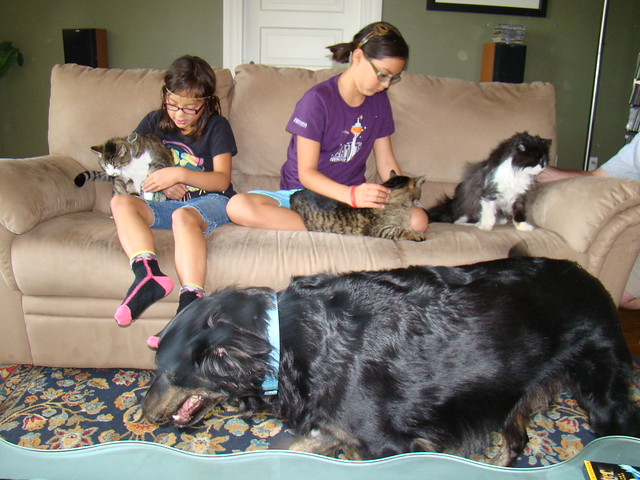}
			\end{tabular}&
			\begin{tabular}{p{4.5cm}}
				Caption: \textit{Some people playing with their pets}
			\end{tabular}
		\end{tabular}
	\end{center}
	\label{fig:snipptets}
	\caption{Examples of the three tasks considered.}
\end{figure}
}


%
%
    %
    %
    %
    %
    %
    %
\section{Related Work}
\label{sec:related_work}


Task-oriented models can be evaluated based on their task success, but
this is not enough to know whether the generated dialogues are
human-like. The development of quantitative metrics to evaluate the
quality of dialogues generated by conversational agents is a 
difficult challenge~\cite{liu-etal-2016-evaluate}, and it is under
investigation for chit-chat dialogue systems. For instance,
Guo et al. \shortcite{guo:topi17} study topic diversity in the conversational flow,
which is rather important in chit-chat and open-domain dialogues, but less so for
task-oriented
ones; Kannan and Vinyals \shortcite{kann:adve16}, Li et
al. \shortcite{li-etal-2017-adversarial}, Bruni and Fern\'andez \shortcite{brun:adve17} propose
to use adversarial evaluation, whereas Lowe et al.
\shortcite{lowe-etal-2017-towards}, See et
al. \shortcite{see-etal-2019-makes}, and Hashimoto et al. \shortcite{tats:unif19} propose
automatic systems that build upon human evaluation. All these efforts
are still preliminary and are not easily employable for new datasets
or new models. Since no standard and unique metric has been proposed to evaluate
the quality of task-oriented (grounded) conversational dialogues, we
consider a mixture of metrics used independently in
various studies, and we provide a comparative analysis
across models and tasks based on the same set of linguistic metrics.



Neural Networks have been shown to generate text that sounds
unnatural due to the presence of repeated utterances, poor vocabulary,
and inconsistency in word usage~\cite{ray-etal-2019-sunny}. Various improvements
have been proposed to mitigate these weaknesses. To
prevent the decoder from choosing words based simply on their frequency, Li et
al. \shortcite{marg:dont19} replace its maximum likelihood estimation
objective, while others change the sampling search
strategy~\cite{holt:cur19,wu-etal-2019-self,see-etal-2019-makes};
these changes aim to reduce the number of repeated questions,
to increase the variety of words and their distribution. Attempts have been made to provide the conversational
models with a reasoning module based on Bayesian
inference~\cite{abba:unce19} or Rational Speech
Act~\cite{sukl:whats19} frameworks that should lead to more
informative and coherent questions.  Here, we do not propose new
models, but rather aim to better understand the strengths and weaknesses
of current models.



\cut{\tb{NNs not good in taking order into account
  \cite{khandelwal-etal-2018-sharp} for texts
  and \cite{wu-etal-2019-self,sank:done19} for dialogues}
\tb{Guo et al., 2018 engagement,
coherence, domain coverage, conversational depth
and topical diversity). Plus attempt to define automatic metrics for chit-chat dialogues of good quality
  \cite{lowe-etal-2017-towards} good idea bu not generalizable
  \cite{see-etal-2019-makes} published?: repetition, specificity,
response-relatedness and question-asking }
\tb{Though PhotoBook: human uses keywords}
\tb{Visual Dialogues:
  \cite{shekhar-etal-2019-beyond} importance to
  evaluate the dialogues also of VS even if they are
  task-oriented. and \cite{ray-etal-2019-sunny} model and metrics that enable quantitative evaluation of consistency in VQA}
\tb{\cite{wu-etal-2019-self,sank:done19} focus on order of questions but 
order of questions in GW human dialogues not that relevant. They are a
collection of info}
\tb{AH AH Dialogue evaluation is relatively well
understood in goal-oriented tasks, where automated approaches can be coded by measuring task
completion}
}

\section{Games and Metrics}
\label{sec:task}

Our focus is on task-oriented dialogues. We consider a task that
relies on grounding language into vision during question generation,
i.e.  GuessWhat \cite{guesswhat_game}, a task that requires
grounding only at the guessing phase, i.e. GuessWhich
\cite{visdial_rl}, and a task based only on language,
i.e. MutualFriends, \cite{he:lear17}.

\begin{table*}
	\centering
	\small
	\begin{tabular}[t]{l|cccccc}\toprule
		& \multicolumn{2}{c}{\#dialogues} & Vocab. size & \#candidates & \#turns & Examples\\
		& training & testing &  &  & &\\\midrule
		GuessWhat & 108K & 23K & 4900 & 3-20& 1-10 & \begin{tabular}{p{4.2cm}}\textbf{\textit{A}}: Is it a person? \textbf{\textit{B}}: No. \\\textbf{\textit{A}}: Is it a dog? \textbf{\textit{B}}: Yes.\\
		{[\textbf{\textit{A}} guesses the target object]} \end{tabular}\\\hline
		GuessWhich & 120K & 2K&  11321 & 2K& 10&\begin{tabular}{p{4.2cm}}\textbf{\textit{A}}: What color is the car? \textbf{\textit{B}}: Blue.\\
		\textbf{\textit{A}}: Who is driving it? \textbf{\textit{B}}: A man.\\ {[\textbf{\textit{A}} guesses the target image]}\end{tabular}\\\hline
		MutualFriends & 8K & 1K & 5325 & 5-12  & 2-46&\begin{tabular}{p{4.2cm}}\textbf{\textit{A}}: My friends work at Google.\\ \textbf{\textit{B}}: None of mine do.\\{[\textbf{\textit{A}} and \textbf{\textit{B}} select a friend]}\end{tabular}\\
		\bottomrule
	\end{tabular}
	\caption{Salient statistics of the human dialogues in the three data sets under consideration in this work. The last column reports samples of dialogues exchanged between two agents (A and B).} \label{tab:dataset-stat}
\end{table*}




\cut{
\begin{table}[t]
	\begin{center}	
	\begin{tabular}{c|c|c}
			{\textbf{GuessWhat?!}} &
			{\textbf{GuessWhich}} &
			{\textbf{Mutual Friends}} 
			\\\hline
			\begin{tabular}{c}
				Is it a person? No\\
				Is it a dog? No\\
				Is it a cat? Yes\\
				The one on the left? Yes
			\end{tabular}
			&
			\begin{tabular}{c}
				How many people are? Two\\
				How old are they? They are young\\
				Is it indoor? Yes\\
				Is there a TV? No
			\end{tabular}
			& 
			\multirow{2}{*}{
				\begin{tabular}{p{4cm}}
					A:Hi! My friends like to do stuff outdoors. \\
					B:Not a single one of mine do! \\
					A:ok so lets go with the indoor friends
					B:Many of my friends work at Google \\
				\end{tabular}
			}
		\end{tabular}
	\end{center}
\end{table}
}
\cut{
\begin{table}
\begin{tabular}{lllll}
& GuessWhat?! & GuessWhich & MutualFriends\\
\#dialogues  & training& 108K & 120K & 8K\\
& testing &   23K  & 2K  & 1K\\
$|$ vocabulary $|$ &  & 4900 & 11324 & 5325\\
\#candidates & & max. 20 & 2K & 8.49\\
\#turns & & 8 & 10 & max. 46\\
\end{tabular}
\end{table}
}

\cut{
\begin{table*}
	\centering
	\small
	\begin{tabular}[t]{l|lllllc}\toprule
		& \multicolumn{2}{c}{\#dialogues} & vocab. & \#candidates & \#turns & example\\
		& training & testing &  &  & &\\\midrule
		GuessWhat?! & 108K & 23K & 4900 & max. 20& 8 & \begin{tabular}{p{3.7cm}}Is it a person? No. \\Is it a dog? Yes.\end{tabular}\\\hline
		GuessWhich & 120K & 2K&  11321 & 2K& 10&\begin{tabular}{p{3.7cm}}Is it red? No, blue. \\Is it daylight? Yes.\end{tabular}\\\hline
		MutualFriends & 8K & 1K & 5325 & 8.49  & < 46&\begin{tabular}{p{3.7cm}}A: My friends work at Google. B: None of mine do.\end{tabular}\\\bottomrule
	\end{tabular}
	\caption{Salient statistics of the three data sets under consideration in this work.} \label{tab:dataset-stat}
\end{table*}
}

\paragraph{Games} As illustrated by the snippets reported in Table
\ref{tab:dataset-stat}, the three tasks also differ in the flexibility
of the dialogues: GuessWhat and GuessWhich are both based on rigid
turns in which an agent asks questions and the other answers, whereas
MutualFriends has free-form dialogues. Moreover, GuessWhat consists
only of Yes/No questions, while in GuessWhich this constraint does not
apply. Relevant statistics of the three datasets are summarized in
Table~\ref{tab:dataset-stat}.
\paragraph{GuessWhat}~\cite{guesswhat_game} is an asymmetric game.\footnote{The dataset of human dialogues is available at
	\url{https://guesswhat.ai/download}.} A
Questioner (Q-Bot) has to ask Yes/No questions to guess which
is the target object among a set of maximum 20 candidates; while asking
questions, it sees the image containing the candidate objects and it
has access to the dialogue history. The Answerer (A-Bot), who knows
which is the target, provides the answers.  The two bots learn to
speak about the image by being trained on human dialogues, which have
been collected by letting humans play the game. Humans could stop
asking questions at any time (human dialogues contain on average 5.2 question-answer pairs), while models have to ask a fixed number
of questions (8 in the setting we have considered). 
\paragraph{GuessWhich}~\cite{visdial_rl} is also an asymmetric
game. Unlike the task described above, the Q-Bot has to ask
questions without seeing the candidate images, but it has access to
captions describing the images. Q-Bot can ask any type of question; the target
image has to be selected among 2K candidates at the end of the
dialogue. The A-Bot instead sees both the caption and the target image. Human
dialogues are from the VisDial dataset\footnote{VisDial is available from
	\url{https://visualdialog.org/data}.} and were collected as chit-chat
dialogues \cite{visdial}. Both humans and models have to ask exactly
10 questions. 
\paragraph{MutualFriends} \cite{he:lear17} is a symmetric game based only
on text: two agents, each given a private list of friends described
by a set of attributes/labels, try to identify their mutual friend
based on the friend's attributes.

\paragraph{Metrics:} Since we are interested in the interplay between
the downstream task and the quality of the generated dialogues, we
consider two types of metrics. 

\textbf{Task Success:} We use the task
success (TS) metrics used in the literature to evaluate models against these
tasks, namely accuracy (ACC) for GuessWhat and MutualFriends, and Mean Percentile Rank (MPR)
for GuessWhich.  The latter is computed from the mean rank position
(MR) of the target image among all the candidates.  An MPR of e.g.,
96\% means that, on average, the target image is closer to the one
chosen by the model than the 96\% of the candidate images. Hence, in
the VisDial test set with 2K candidates, 96\% MPR corresponds to an MR
of 80, and a difference of $\pm$ 1\% MPR corresponds to $\mp$ 20 mean
rank. The task success chance levels are: 5\% accuracy (GuessWhat), 50\% MPR
  (GuessWhich) and 11.76\% accuracy (MutualFriends).

\textbf{Linguistic metrics:} 
It has been shown that the quality of the dialogues generated by computational agents is not satisfactory. The main weaknesses of these models consist of poor lexical diversity, a high number of repetitions, and the use of a limited vocabulary.  To evaluate the quality of the generated
dialogues (defined as the closeness to human dialogues according to surface-level cues), we use several metrics that have been proposed in the
literature. As in He et al. \shortcite{he:lear17}, we compute \emph{unigram
  entropy (H)}, which measures the entropy of
unique unigrams in the generated dialogues normalized by the total number of tokens used by the model. From Murahari et al. \shortcite{mura:impr19}, we take the \emph{Mutual Overlap (MO)} metric, which evaluates the
 question diversity within a dialogue by computing the average of the
 BLEU-4 score obtained by comparing each question with the other
 questions within the same dialogue.\footnote{A high number of novel questions
 	and low mutual overlap cannot be taken per se as a sign of high
 	quality of the dialogues: a model could ask a question never seen in
 	training or with very little overlap with the other questions but
 	completely out of scope. To rule out this possibility, we compute the cosine similarity of each question
 	marked as novel and with a low mutual overlap with the
 	dialogue they occur in, and compare it with the similarity
 	between the latter and random questions taken from other
 	dialogues. Embeddings are obtained by using Universal Sentence Encoder-USE
 	\cite{cer:univ18}. We found that novel and low-MO questions are more similar to their dialogue than the
 	random ones, confirming the effectiveness of these metrics.}
Moreover, following~\citet{shekhar-etal-2019-beyond}, we report the percentage of games with \emph{one question
repeated verbatim (GRQ)} within a dialogue. Finally, we
compare models with respect to their ability on lexical acquisition by
calculating the \emph{Global Recall (GR)} introduced by 
\citet{van-miltenburg-etal-2018-measuring} to evaluate image
captioning: it is defined as the overall percentage of learnable words
(from the training set) that the models recall (use) during
generation. Furthermore, taking inspiration from the Local Recall introduced in the
same work, we propose a similar metric tailored to dialogues, i.e.,
\emph{Local Recall-d (LRd)}, which measures  how many content words the generated dialogue shares with the corresponding human dialogue for the same game.  Given a human dialogue $D_h$ about an
image and a generated dialogue $D_g$ about the same image, we compute
LRd as the normalized lexical overlap (considering only content words) between
$D_h$ and $D_g$. 

We sum up all these linguistic metrics used in the literature so far
into one which we take as a proxy of the quality of dialogues: it shows the
\emph{linguistic divergence (LD)} of the dialogues generated by a
model from human dialogues. To this end, we normalize each metric so that all values lie between 0 and 1: 0 stands for human performance for the ``lower is
better'' metrics and 1 stands for human performance for
``higher is better'' metrics. We compute LD by
averaging all the scaled values V for each model; we take $1-V$ for
``higher is better'' metrics to obtain a ``divergence'' value. All
  the metrics are equally weighted. By definition, LD is 0 for human
  dialogues. LD captures three main surface-level aspects: overall vocabulary usage
(H, GR), diversity of questions/phrases within a dialogue (MO, GRQ),
and similarity of content word usage with respect to human dialogues
(LRd). There could be some correlation between metrics capturing
similar aspects of language quality, but this does not affect the
validity of the proposed LD metric.




\section{Models}
\label{sec:models}

For both visual dialogue games, GuessWhat and GuessWhich, supervised
learning has been compared with other learning paradigms.  After
the introduction of the supervised baseline model
\cite{guesswhat_game}, several models have been proposed for
GuessWhat. They exploit either reinforcement learning 
\cite{sang:larg18,zhan:aski18,zhang:mult18,zhao:impr18,gan:mult19,pang:visd20} or cooperative
learning  \cite{shekhar-etal-2019-beyond,pang:visd20}; in both
cases, the model is first trained with the supervised learning regime
and then the new paradigm is applied. This two-step process has been
shown to reach higher task success than the supervised approach. For
GuessWhich, after the supervised model introduced in \citet{visdial}, new models based on reinforcement learning
have been proposed, too~\cite{visdial_rl,mura:impr19,zhou:buil19}, but
their task success is comparable if not lower than the one achieved  by
using only supervised learning (see \citealt{test:thed19}). Below, we briefly describe the models we
have compared in our analysis. For each task, we have chosen
generative models trained with different learning paradigms and for
which the code is available; for each paradigm, we have tried to choose the
best performing ones or those that obtain a task success near to
state-of-the art and could help better understand the interplay
between task success and dialogue quality.


\paragraph{GuessWhat}  We use the A-Bot introduced
in de Vries et al. \shortcite{guesswhat_game}, which is trained in a supervised learning (SL) fashion.  
For the Q-Bot, we compare models based on
different learning paradigms: supervised and cooperative learning
(GDSE-SL and GDSE-CL, respectively) proposed in~\citet{shekhar-etal-2019-beyond} and
reinforcement learning (RL) proposed in~\citet{stru:end17}. In RL,
the reinforce paradigm used aims at optimizing the task accuracy 
of the game. Besides
using different learning paradigms, these models differ in their
architecture. In particular, while in RL the Question Generator (QGen)
and the Guesser are trained independently, in GDSE a common
visually-grounded dialogue state encoder is used and the two modules
are trained jointly. In both cases, the Guesser receives as input the
candidate object's categories and their spatial coordinates, and
  during training it
is updated only at the end of the dialogue.\footnote{The code of the A-Bot and of RL is
  available at \url{https://github.com/GuessWhatGame/guesswhat}. The code of GDSE at:
\url{https://github.com/shekharRavi/Beyond-Task-Success-NAACL2019}.} 

\paragraph{GuessWhich}
We use the A-Bot introduced in \citet{visdial_rl}. 
For the Q-bot, we compare Diverse \cite{mura:impr19} and ReCap \cite{test:thed19}.
Diverse and ReCap have similar architectures: several
encoders incrementally process the linguistic inputs to produce
the dialogue hidden state. This state is used to condition a decoder
that generates a new question at each turn, and a guesser that is
trained to produce the visual representation of the target image through a feature regression module. The
two models differ in the encoders used and in the training
paradigm. While Diverse encodes the caption together with the dialogue history through a Hierarchical LSTM, ReCap has two independent LSTMs that produce the linguistic features of the caption
and of the dialogue history, merged together to
produce the dialogue hidden state. 
Secondly, in Diverse, an auxiliary objective
on the dialogue state embedding (Huber loss) is used to incentivize the bot to ask more diverse questions with respect to
 the immediate previous turn.\@
\cut{First of all, in Diverse, at each turn an LSTM encodes the new
question-answer pair whose embedding is given as input to a
Hierarchical LSTM that processes it together with the caption and the
dialogue history. Instead, ReCap re-reads the caption at each turn:
two independent LSTMs produce the linguistic features of the caption
and of the dialogue history which are merged by a dialogue encoder to
produce the dialogue hidden state. Secondly, in Diverse, the decoder and the
Guesser are updated at every turn and an auxiliary objective (Huber loss)
on the dialogue state embedding is used to incentivize the bot to ask
diverse questions.
The model is penalized whenever it asks a
follow-up question similar to the one of the immediate previous turn.}
In ReCap, the Guesser sees the
ground-truth image only at the end of the game while in Diverse the Guesser is updated at each turn.  ReCap has been
trained only by SL, Diverse both by SL (D-SL) and SL plus RL (D-RL). Further details can be found in the respective
papers \cite{mura:impr19,test:thed19}.\footnote{The code for the
  A-Bot model and
 for D-SL and D-RL is available at
  \url{https://github.com/vmurahari3/visdial-diversity};  it is not
  specified how the best models are chosen. For ReCap:
 we have obtained the code by the authors and trained the model;
 we have chosen the model whose MPR does not increase for the subsequent 5 epochs.}

\cut{{\tt Diverse} has a similar structure to the {\tt A-bot}
described above but it is also trained to obtain a visual
representation. It consists of an encoder that
receives first the caption and then the question-answer pairs
sequentially; it outputs a state-embedding at round $t$ that is
jointly used by the decoder and by a Feature Regression Network
(FRN). The decoder and the FRN are updated at every turn and an auxiliary objective (Huber loss) on the dialogue state
embedding is used to incentivize the bot to ask diverse questions.  The model
is penalized whenever it asks a follow-up question similar to the one
of the immediate previous turn.  
We report the results obtained
by the model when trained only by SL  ({\tt
  D-SL}) or by a SL plus  RL  ({\tt D-RL}).\footnote{The best performing -SL and -RL
  models are released in the authors' github
  \url{https://github.com/vmurahari3/visdial-diversity}.}  The overall
architecture of {\tt ReCap} is rather similar to the one of {\tt
  Diverse} but it re-reads the caption at each turn.  Two independent
LSTMs produce the linguistic features of the caption and of
the dialogue history. These two representations are passed to a
dialogue encoder and the final layer is given as input to both the
question decoder (QGen) and the Guesser.  QGen employs an LSTM
network to generate the token sequence for each question. The Guesser
acts as the FRN of  {\tt Diverse}: it takes as input
the dialogue hidden state produced by the dialogue encoder, and passes
it through two linear layers with a ReLU activation function on the
first layer. In
contrast to the FRN, the
Guesser sees the ground-truth image only at the end of the game.
The model is trained with SL.\footnote{We have obtained the code by the authors.}
}



 
\cut{The {\tt Diverse}
model has a similar structure to the {\tt A-bot} described above and
it is trained to obtain a visual representation by a regression
module. It consists of an encoder that receives first the caption and
then the question-answer pairs sequentially; it outputs a
state-embedding at round $t$ that is jointly used by the decoder (an
LSTM which learns to generate the next question) and by a Feature
Regression Network (FRN, a fully connected layer which learns to
approximate the visual vector of the target image). The decoder and
the FRN are updated at every turn.  Differently from the {\tt Q-Bot}
of~\cite{visdial_rl}, the Diverse model is trained with an auxiliary
objective (Huber loss) on the dialogue state embedding that incentivizes
the bot to ask diverse questions. \at{Repetition - we already mention
  this above.} The model is penalized whenever it asks a follow-up
question similar to the one asked in the immediate previous turn.  We
report the results obtained by the model when trained only by
supervised learning ({\tt Diverse-SL}) or by a SL phase followed by a
Reinforcement Learning (RL) one ({\tt Diverse-RL}).\footnote{The best
  performing -SL and -RL models are released in the authors' github
  \url{https://github.com/vmurahari3/visdial-diversity}.}

\paragraph{ReCap} The overall architecture of the {\tt ReCap} model
proposed in~\cite{test:thed19} is similar to the one of the {\tt
  Diverse} model. Crucially, it re-reads the caption at each turn. 
Two independent LSTM networks produces the linguistic features of the
caption and of the dialogue history. These two representations are passed to
a dialogue encoder: they are concatenated and scaled through a linear layer
with a \emph{tanh} activation function. The final layer (viz.\ the
dialogue state) is given as input to both the question decoder (QGen)
and the Guesser module.  QGen employs an LSTM network to generate the
token sequence for each question. The Guesser module acts as the
feature regression network (FRN of the {\tt Diverse} model): it takes
as input the dialogue hidden state produced by the dialogue encoder, and passes
it through two linear layers with a ReLU activation function on the
first layer. The final representation is a 4096-d vector which
corresponds to the fc7 VGG representation of the target image. In
contrast to the FRN by~\cite{visdial_rl}, as mentioned above, the
Guesser ``sees'' the ground-truth image only at the end of the game.
The model is trained in a  Supervised Learning fashion.\footnote{\rb{The
code is released in the authors' github \url{https://vista-unitn-uva.github.io/}}.} \at{We should say that Recap architecture is similar to Ravi NAACL-2019 + we can say that we refer to the Testoni et al. and Das et al. for the models' picture.}
}

\paragraph{MutualFriends} We evaluate the model proposed
in \citet{he:lear17}, DynoNet (Dynamic Knowledge Graph Network), in
which entities are structured as a knowledge graph and the utterance
generation is driven by an attention mechanism over the node
embeddings of such graph. The model is trained via supervised learning
and at test time it plays with itself. DynoNet consists of three
components: a dynamic knowledge graph (which represents the agent's
private KB and shared dialogue history as a graph), and two LSTMs that map the graph embedding over the nodes and generate
utterances or guess the entity.\footnote{The code is available at
  \url{https://github.com/stanfordnlp/cocoa/tree/mutualfriends}.}

\section{Experiments and Results}
\label{sec:experiments}

\citet{shekhar-etal-2019-beyond} has shown that in GuessWhat task
success (TS) does not correlate with the quality of the dialogues. First of
all, we check to what extent this result is task and model agnostic
by taking GuessWhat, GuessWhich and MutualFriends as case-studies
and compare the behaviour of the models described above.  

First, we
evaluate the impact of the number of epochs and the size of the training set;
then, we study whether and when a lower LD could help
to reach a higher TS.

\subsection{Task Success and Linguistic Divergence}

We evaluate all models described above, in their supervised,
cooperative, or reinforcement learning version, aiming to test whether
some patterns can be found irrespectively of the model and data
explored.  Our results confirm what has been shown
in~\citet{shekhar-etal-2019-beyond} for GuessWhat: \textbf{TS does
  not correlate with the quality of the generated dialogues}; models with similar TS
generate dialogues that vary greatly with respect to the
linguistic metrics. We run a Spearman's analysis and found a very weak
correlation between LD and TS (coefficient < 0.15, p-value <
0.05). 
Our comparison across tasks shows that both in
GuessWhich and in GuessWhat the vocabulary used by humans while
playing the games in the training and testing set is rather similar
(resp., 91\% and 84\% of words are in common between training and
testing sets). Yet in both visual tasks models reach an LRd of
  around 42\%. Specifically, the
  average mean rank of words they fail to use is 7000 (over 11321) for
  GuessWhich and 3016 (over 4900) for GuessWhat. \textbf{Hence, models mostly use very frequent words.} Details on the metrics for each task and model are reported in Table \ref{tab_res}.

\begin{table*}[!ht]
	\centering
	\resizebox{\textwidth}{!}{%
		\begin{tabular}{|c|c|c|c|c|c|c|c|c|c|c|}
			\hline
			\multicolumn{1}{|l|}{} & \multicolumn{4}{c|}{\textbf{GuessWhich}}         & \multicolumn{4}{c|}{\textbf{GuessWhat}}           & \multicolumn{2}{c|}{\textbf{MutualFriends}}                           \\ \hline
			\multicolumn{1}{|l|}{}          & D-SL  & D-RL  & ReCap-SL & Hum                                           & GDSE-SL & GDSE-CL & RL    & Hum                                           & DynoNet-SL                                    & H                                             \\ \hline
			\textbf{TS $\uparrow$}                                           & 95.2  & 94.89 & 96.76    & - & 48.21   & 59.14   & 56.3  & 84.62                                         & 0.98                                          & 0.82                                          \\ \hline
			\textbf{GR $\uparrow$}                                           & 6.46  & 9.04  & 14.4     & 27.69                                         & 34.73   & 36.35   & 12.67 & 72.98                                         & 51.15                                         & 65.2                                          \\ 
			\textbf{LRd $\uparrow$}                                          & 39.93 & 41.83 & 42.76    & - & 42.1    & 42.41   & 34.51 &- & - & - \\ 
			\textbf{MO $\downarrow$}                                           & 0.51  & 0.41  & 0.23     & 0.07                                          & 0.39    & 0.23    & 0.46  & 0.03                                          & - & - \\ 
			\textbf{GRQ $\downarrow$}                                          & 93.01 & 81.17 & 55.37    & 0.78                                          & 64.96   & 36.79   & 96.54 & 0.8                                           & - & - \\
			\textbf{H $\uparrow$}                                            & 4.03  & 3.92  & 4.19     & 4.55                                          & 3.52    & 3.66    & 2.42  & 4.21                                          & 3.91                                          & 4.57                                          \\ 
			
			\textbf{LD $\downarrow$}                                       & 0.58  & 0.52  & 0.38     & -                                             & 0.46    & 0.36    & 0.67  & -                                             & 0.18                                          & -                                             \\ \hline
	\end{tabular}}
	\caption{Comparative analysis of different models on several tasks and datasets.TS: task success. GR: global recall. LRd: local recall. MO: mutual overlap. GRQ: games with repeated questions. H: unigram entropy. LD: linguistic divergence. $\uparrow$: higher is better. $\downarrow$: lower is better.}
	\label{tab_res}
\end{table*}


%

\begin{figure}[t]
	\centering \hspace*{-0.1cm}
	\includegraphics[width=1\linewidth]{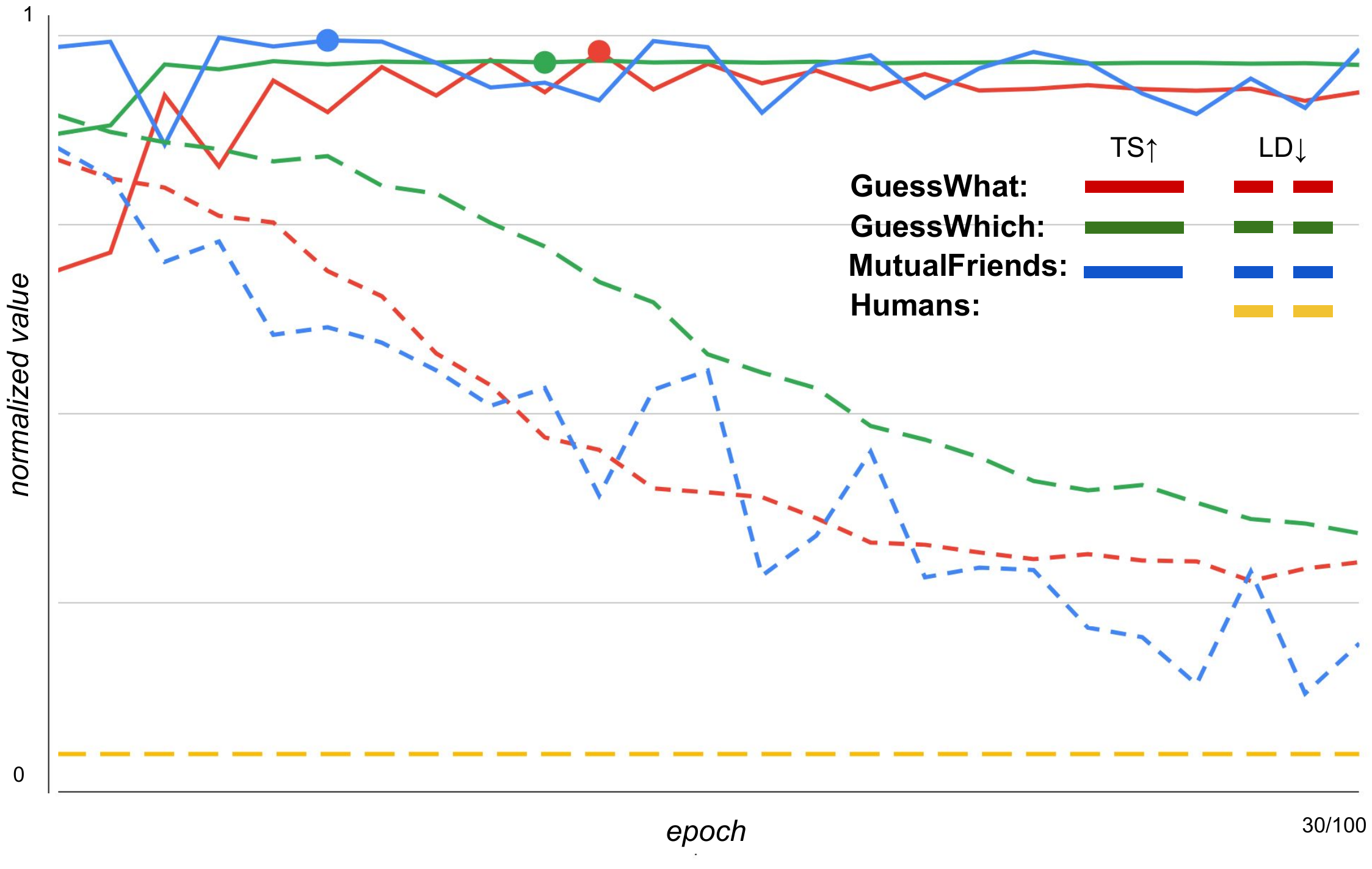}
	\caption{Comparison of Task Success (TS, solid lines) and Linguistic
		Divergence (LD, dashed lines) for GDSE-SL (GuessWhat, trained for 100 epochs), ReCap (GuessWhich, trained for 100 epochs),
		DynoNet (MutualFriends, trained for 30 epochs); Humans LD lower-bound metric in
		yellow. The LD of the generated dialogues keeps decreasing (moving
		close to human level) even though we no longer notice
                improvements in TS, whose highest value is reached
                well before (marked by bullets). } 
	\label{fig:summary}
\end{figure}

\subsection{Learning Processes behind TS and LD}
We aim to understand the relation between TS and LD.  To this end, we
compare the two metrics during the training processes across epochs and by downsizing the
training set. For each task, we consider the models trained in a
SL fashion since those trained with other paradigms
build on them. Hence, we focus on ReCap, GDSE-SL, and DynoNet.

\begin{figure*}[t]\centering 
	\includegraphics[width=1\linewidth]{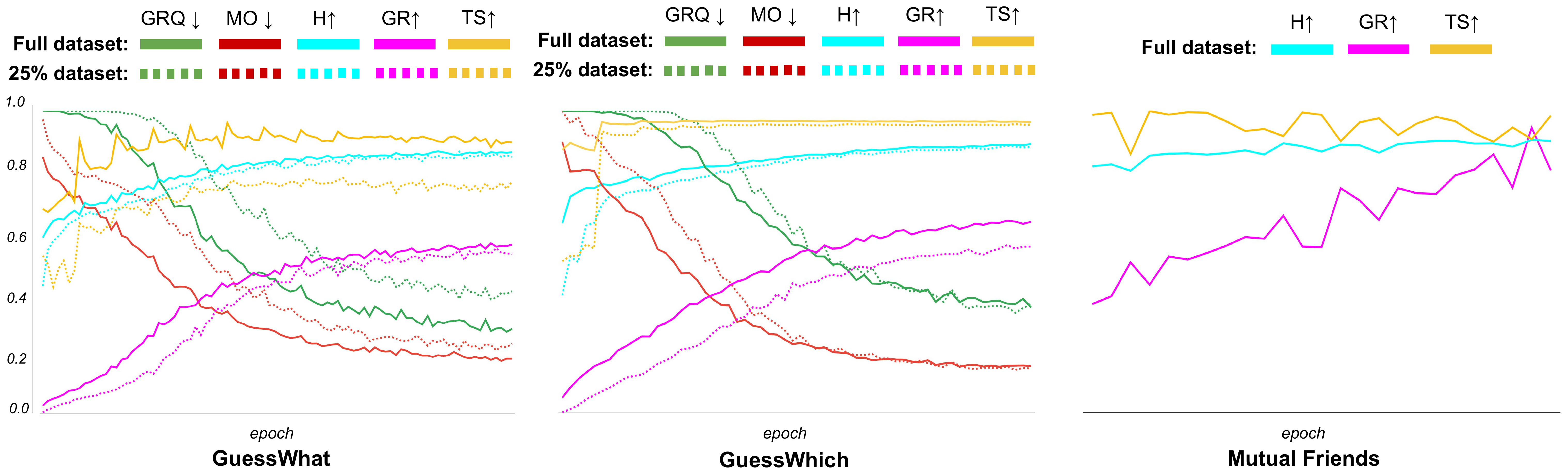} 
	\caption{Comparison across epochs and by downsizing the
		training data using the following metrics: Task
		Success (TS), Games with Repeated Questions (GRQ), Mutual
		Overlap (MO), Unigram Entropy (H) and Global Recall (GR);
		all metrics are scaled between 0 and 1 (y-axis).
		\textbf{Left:} GDSE-SL on GuessWhat. \textbf{Middle}:
		ReCap on GuessWhich.  \textbf{Right}: DynoNet on
		MutualFriends. Downsizing the training data has a higher
		impact, both for TS and some linguistic metrics, in
		GuessWhat than in GuessWhich. Among the linguistic
		metrics, entropy is the most stable and GR increases through
		the epochs in all tasks. For readability, we have not reported Local Recall-d
		since its pattern is very close to GR.  The dataset
		of MutualFriends is too small to analyse the effect of
		downsizing it.} \label{fig:epochs-datasize} 
\end{figure*}

\paragraph{Comparison Across Epochs}
We study  for \emph{how long} a model has to be trained to
reach its best performance on guessing the target referent and
generating human-like dialogues.
Figure~\ref{fig:summary} reports the TS and the LD of three models trained on the three tasks under examination. Each line is normalized w.r.t the highest value for each metric, so that it is possible to see different trends on the same plot.
As we can see from the figure, the highest TS (marked by bullets) is reached
earlier in GuessWhich and in MutualFriends than in GuessWhat.
More interestingly, for all the tasks, the LD of the generated
dialogues keeps decreasing (moving close to human level) even though
we no longer notice improvements in TS, whose highest value is reached
well before.  Figure~\ref{fig:epochs-datasize} (solid lines) reports the details of
the linguistic metrics used to compute LD. We see that for all  tasks
a high entropy is reached already after a few epochs; this means that
though the number of words used is small, models learn to distribute
their use well. All the other metrics improve through the
epochs quite a lot. For MutualFriends, we do not compute MO and GRQ
since the model trained on it, DynoNet,  asks questions referring to different attributes
and hence, by design, it generates very few repetitions. 
From the results of this first experiment, it
emerges fairly clearly that in all referential games we have
  considered, \textbf{models learn to perform well on the task quite
  quickly.} On the one hand, this means that \textbf{choosing the best model
  purely on the basis of its TS prevents the model from developing better
  linguistic skills}, on the other hand, that \textbf{the higher quality of the
  dialogues does not help reach a higher TS.} This result holds in
all cases despite the target being an entity in a graph described by
linguistic attributes (MutualFriends), an object (GuessWhat) or an
image (GuessWhich).


\paragraph{Comparison by Downsizing the  Training Set}
To understand whether the relation between TS and LD is related to the
size of the training set, we compare models trained on datasets
of decreasing size. We evaluate the models by training them with
50\% and 25\% of the standard GuessWhat and GuessWhich datasets. For
MutualFriends, we have not run the downsizing analysis since the
dataset is too small.  For readability reasons, in
Figure~\ref{fig:epochs-datasize} we report only the results obtained
with the 25\% setting since they represent the observed pattern well
enough. The y-axis reports the metrics scaled between 0 and 1.
In GuessWhich the TS (yellow lines) does not decrease by downsizing
the dataset: when using just 25\% of the full dataset (dotted
line) it gets very close to the highest MPR obtained by the model
trained on the full dataset (solid line) already after the first 5
epochs. Interestingly, the linguistic metrics do not get worse either, with
the only exception of GR. However,  in GuessWhat the TS
decreases when downsizing the training data (again, yellow solid
vs.\ dotted lines) and dialogues quality is affected too (with the
exception of entropy and GR). This result shows that \textbf{in
  GuessWhat \textit{how well} the model learns to ground language plays an important role and affects the TS.} 
In the next experiment, we aim to further understand the difference
between the two visual tasks and when LD could impact TS in GuessWhat.




\subsection{When Could Language Quality Impact Task Success?}
First of all, we check the extent to which the dialogue is used by the
Guesser module to pick the correct target.  Secondly, we evaluate whether the quality
of the dialogues could lead to higher task success. Finally, we
pinpoint when a lower LD could contribute to succeed in the task.

\paragraph{The role of the dialogues on TS}
We run a by-turn evaluation checking whether the information
incrementally gained through the dialogue brings increased
performance. We evaluate ReCap on GuessWhich and GDSE-SL on GuessWhat.
We find that the performance of ReCap is flat across the dialogue
turns, confirming results reported in~\citet{mura:impr19} for other
models. Instead, the performance of GDSE-SL keeps on increasing at
each turn from the beginning till the end of the dialogue, though the
increase in the first 3 turns is higher than in the later
ones (details in Appendix \ref{sec:appendix_A}). \textbf{This suggests that in GuessWhich
  the role of the dialogue is rather limited.} This might be due to the highly informative image caption that GuessWhich
models receive together with the dialogue to solve the guessing
task~\citep{test:thed19}.
Instead, in \textbf{GuessWhat dialogues do play a major role in the
  guessing task}. Hence, we focus on this dataset to understand
whether and when the quality of the dialogue could lead to a higher
task success.

\begin{figure*}[t]
	\centering
	\begin{minipage}{4cm}
		\includegraphics[width=4cm]{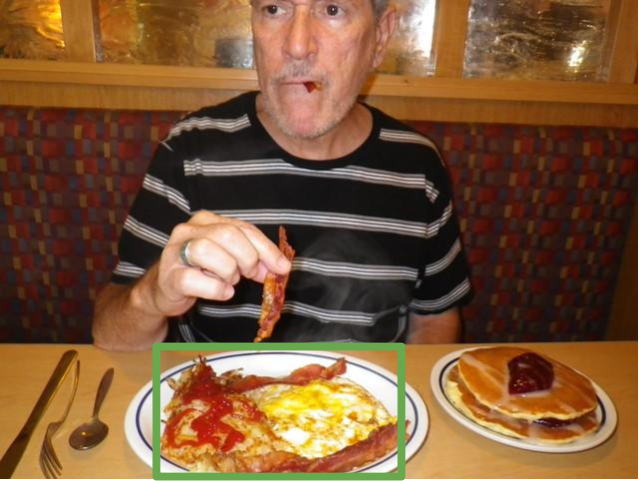} 
	\end{minipage}
	\begin{minipage}{9cm}\small
		\begin{tabular}[t]{l@{\hspace{-0.2cm}}r|l@{\hspace{-0.2cm}}r}
			\multicolumn{2}{c}{Human dialogues} & \multicolumn{2}{c}{Generated dialogues}\\
			\bf Questioner & \bf Answerer & \bf Questioner & \bf Answerer\\[3pt]
			1. Is it a man? & No & 1. Is it a person? & No\\[3pt]
			2. Is it food?  & Yes & 2. Is it food? & Yes\\[3pt]
			3. Is \emph{pancake}?  & No & 3. Is it pizza? & Yes \\[3pt]
			4. Is egg? & Yes & 4. Is it the pizza in front? & Yes\\[3pt]
			&  & 5. \ldots & \\
			\multicolumn{2}{l}{\bf $\leadsto$ model suceeds guessing} &
			\multicolumn{2}{l}{\bf
				$\leadsto$
				model
				fails guessing} \\
		\end{tabular}
	\end{minipage}
	\vspace*{.6cm}
	
	\begin{minipage}{4cm}
		\includegraphics[width=4cm]{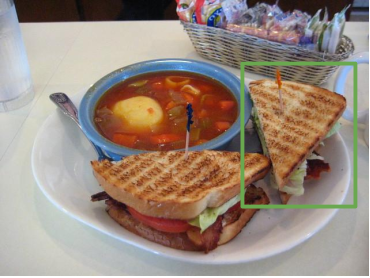} 
	\end{minipage}
	\begin{minipage}{9cm}\small
		\begin{tabular}[t]{l@{\hspace{-0.2cm}}r|l@{\hspace{-0.2cm}}r}
			\multicolumn{2}{c}{Human dialogues} & \multicolumn{2}{c}{Generated dialogues}\\
			\bf Questioner & \bf Answerer & \bf Questioner & \bf Answerer\\[3pt]
			1.Is it edible? & Yes & 1. Is it food? & Yes\\[3pt]
			2.Is it a sandwich? & Yes & 2. Is it a sandwich? & Yes\\[3pt]
			3. Does it have an orange & && \\ \textit{toothpick} in it?  & Yes & 3. Is it on the right? & Yes \\[3pt]
			 & & \ldots &\\
			\multicolumn{2}{l}{\bf $\leadsto$ model fails guessing} &
			\multicolumn{2}{l}{\bf
				$\leadsto$
				model
				succeeds guessing} \\
		\end{tabular}
	\end{minipage}\caption{Examples of GuessWhat games in which
          humans use ``rare'' words (rare words in italic) and the
          corresponding generated dialogues. The failure of the
            model could be due to the inability to generate (top) or  to
            encode (bottom) rare words. }\label{fig:sample}
	\vspace*{-5pt}
\end{figure*}

\begin{figure}[t]\centering 
	\begin{tabular}{l}
		\includegraphics[width=1\linewidth]{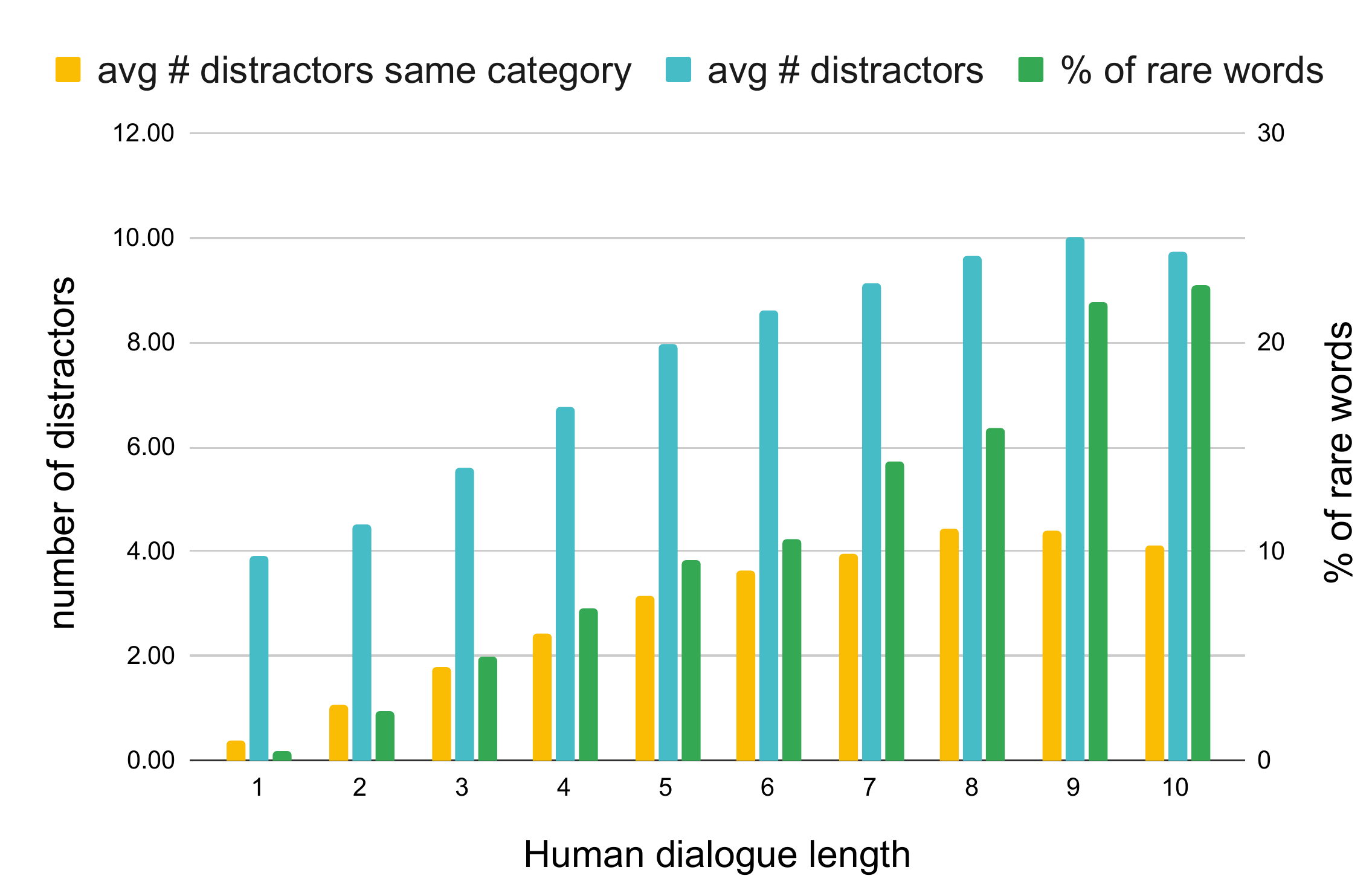}
	\end{tabular}
	\caption{In GuessWhat, longer human dialogues contain more
		rare words, more distractors and
		more distractors of the same category of the target object.} 
	\label{fig:lenght}
\end{figure}

\paragraph{Impact of the quality of dialogues on TS}
To check whether the Guesser could profit from dialogues of better
quality, we evaluate GDSE-SL using human dialogues. When given human
dialogues, the model reaches an accuracy of 60.6\%, which is +8.5\%
higher than the one it achieves with the dialogues generated by its
decoder (52.1\%).
One hypothesis could be that this higher TS is due to the mistakes produced by the A-bot when using instead the generated dialogues,
but this is not the case: we have evaluated the model when receiving
human questions paired with the A-bot's answers for each question and the accuracy drops
of only 2.5\%. This experiment suggests that \textbf{a lower LD could indeed lead to a
higher TS.}


\paragraph{The role of less frequent words}
As we have observed above, models mostly use very frequent words. 
Here, we aim to understand to what extent this penalizes GuessWhat models. In this
dataset, more than half (55\%) of the words in the vocabulary are used less than 15 times in the training set. We refer to this set of words as ``\emph{rare}'' words: most of them are nouns (79\%) or verbs (11\%) (e.g.,
``\emph{feline}'', ``\emph{forest}'', ``\emph{compute}'',
``\emph{highlight}'').

We check whether there is a relation between
rare words and difficult games. Human dialogue length is a good proxy
of the difficulty of the games, both for humans and models.  Figure~\ref{fig:lenght} illustrates some statistics about
human dialogues: games for which humans ask more
questions are about images with a higher number of distractors, with a
higher number of distractors of the same category of the target, and
with a higher number of ``rare'' words.  In 10\% of the games in the
test set, humans have used at least one rare word. These dialogues are
longer than those that do not contain rare words (resp., 7.8 vs. 4.7
turns on average).  Interestingly, the accuracy of the model on these
games is lower than the overall accuracy: -13.8\% (48.7\% vs. 62.5\%)
when evaluating it with human dialogues and -8.3\% (45\% vs. 53.3\%)
when using the dialogues generated by the model itself. Moreover,
the accuracy reached by the model in the latter setting is
lower when comparing games for which humans have used a higher number
of rare words. Overall, we found that 65\% of the rare words in the human test set show up in games that the model is not able to solve correctly.

Figure~\ref{fig:sample} shows some examples of games
in which humans have used a rare word. It illustrates the  human
vs.\ generated dialogue and whether the model succeeds in guessing the target object
when receiving the former or the latter. The failure of the
  model in guessing the target object could be due to its inability to
generate or encode rare words. The example on top shows that
if the model fails to generate an appropriate word (e.g.\ the rare word
``\emph{pancake}'') this can have a domino effect on the next words
and the next questions it generates. On the other hand, the model can fail
to encode rare words, e.g., ``\emph{toothpick}'' in
Figure~\ref{fig:sample}-bottom. The inability to generate rare words
could be mitigated by developing dialogue strategies that produce
less natural but still
informative dialogues. For instance, in the example at the
bottom, the model avoids using
``\emph{toothpick}'' by asking a spatial question (``\textit{Is it on
  the right?}'') which is rather
informative for the Guesser since it has the coordinates of each
candidate object.
\cut{In this case, it is worth pointing out that the generated
  dialogue exploits the strengths of the current architecture for
  succeeding in the game by asking a question about the position of
  the object in the image (``\textit{Is it on the right?}'') without
  using rare words; the current Guesser architecture, in fact,
  receives the category and the spatial coordinates of each candidate
  object.} 
These observations show that current models fail to properly ground
and use \emph{rare words} and suggest that, in some contexts,
  \textbf{the use of only frequent words could be behind the failure in reaching the
  communication goal.} 

\cut{
  We acknowledge that the model could avoid using rare words to succeed in the game by exploiting some biases in the dataset. For instance,  referring to the position of the objects in the image may allow models to achieve a good task success without using rare words. Indeed, models trained according to a Reinforcement Learning paradigm show this behaviour: they generate utterances containing mostly (if not only) spatial questions. Unfortunately, the dialogues thus generated sound unnatural and highly repetitive, while we are seeking for models that generate human-like utterances. 
}

\cut{NOT USED: By using data perturbation, it has been shown that neural networks
(NNs) language models are
capable of using context consisting of about 200 tokens on average, but
they have more difficulty in modelling tokens beyond the first nearest
50th positions~\cite{khandelwal-etal-2018-sharp}. By applying a
similar analysis, Sankar et al. \shortcite{sank:done19} have shown
that NN conversational agents are not able to take the dialogue
history into account. We wonder whether this can be the reason behind
the failure of the models to exploit a better language proficiency.
}

\cut{NOT USED: In the literature on GuessWhat?! and GuessWhich, attention has been
so far put only on the Q-Bot; as a consequence, the
A-Bot is usually frozen through the epochs. Wrong answers by the A-Bot could provide even
inconsistent information as shown in Li et
al. \shortcite{marg:dont19}; Ray et
al. \shortcite{ray-etal-2019-sunny}. Hence, the mistakes coming from the A-Bot could
prevent the Q-Bot reaching a higher performance. For instance, the
Oracle commonly used for GuessWhat?! has an accuracy of around 80\%,
which means there is a 83\% probability that an 8-turn generated
dialogue contains at least one wrong answer, but not much  is known
about its impact. Similarly, the TS could be influenced by the
model having asked ``wrong'' questions.

To better understand the impact of the mistakes made by the decoder or
by the A-bot, we
have evaluated ReCap and GDSE-SL across the epochs on the dialogues asked
by humans for test set games. The same TS pattern has emerged: the
models reach a high TS pretty quickly (around epoch 30 for GDSE-SL and
around epoch 25 for ReCap). \textbf{Result} This suggests that the difficulty of the
models to exploit an higher linguistic proficiency most probabily \emph{is not} due to the
mistakes made by the A-bot nor to the failure of the decoder in
generating good questions. 
}

\section{Conclusion}
\label{sec:conclusion}
Our work highlights the different complexity of two sub-tasks involved
in referential guessing (visual) games: guessing the target and asking
questions. We have shown that while learning to win the game can be
reached in a rather short time (with a few epochs), learning to
generate rich human-like dialogues takes much longer. This holds for
all three tasks and models we have scrutinized independently of the
size of the vocabulary, the task, and the learning paradigm
used. Therefore, choosing the best model only on the base of the
task success could prevent the model from generating more human-like
dialogues.  We have shown
that in GuessWhich decreasing the size of the training set does not bring a drop in
either TS (task success, higher is better) or in LD (linguistic divergence, lower is better) and, moreover, the dialogues play a minor role on
TS.  Instead, for GuessWhat, decreasing the size of the training
dataset brings a decrease in TS and an increase in LD, and, through dialogues,
models accumulate information to succeed in the task. Hence, we have
focused our in-depth analysis on GuessWhat. Furthermore, we have investigated whether and when higher
language quality could lead to higher task success. We have shown that if
models are given human dialogues, they can reach a higher
TS. Hence, LD could boost TS. We have shown that this boost could help more in difficult games, i.e. those for which humans ask longer
dialogues. These games contain images with more distractors and humans
use less frequent words while playing them. Hence, we claim that in GuessWhat models
could increase their accuracy if they learn to ground, encode and
decode words that do not occur frequently occur in the training set. 

In the paper, we propose the LD metric that, despite its limitations (i.e., being based only on surface cues) represents a proxy of the quality of dialogues. We believe LD effectively captures the most common deficiencies of current models and it allows a straightforward comparison between different models. As future work, LD can be used as a training signal to improve the quality of generated dialogues. Moreover, a comparison between human quality judgments and LD may shed some light on the strengths and weaknesses of this metric. Further work is needed to design new metrics that capture more fine-grained phenomena and better evaluate the quality of generated dialogues.

\cut{ we have shown that in GuessWhich models
can be successful in learning both skills from a quite small training
data set, while in GuessWhat?! the performance of models and the
quality of the dialogues decrease when reducing the dataset.  We
conjectured that the reason for such discrepancy between the two
learning curves is due to the encoder failing to handle the dialogue
history or to the mistakes made by either the decoder or the Oracle,
but our results do not confirm these hypotheses.  We have touched up
an issue overlooked in visual dialogues, namely the impact of ``rare
words''. Our preliminary analysis suggests that they could affect both
task success and the quality of dialogues. We believe the issue requires
a more in-depth investigation and might speak in favour of using
pre-trained encoder-decoder models.
}






\section*{Acknowledgements}
The authors kindly acknowledge the support of NVIDIA Corporation with the donation of the GPUs used in our research. We are grateful to SAP for supporting the work. We would like to thank the following people for their suggestions and comments: Luciana Benotti, Guillem Collell, Stella Frank, Claudio Greco, Aurelie Herbelot, Sandro Pezzelle, and Barbara Plank. Finally, we thank the anonymous reviewers for the insightful feedback.


\bibliographystyle{acl_natbib}
\bibliography{../raffa}

\newpage
\appendix

\section{Appendix A}
\label{sec:appendix_A}
\begin{figure}[b!]
	\centering 
	\includegraphics[width=1\linewidth]{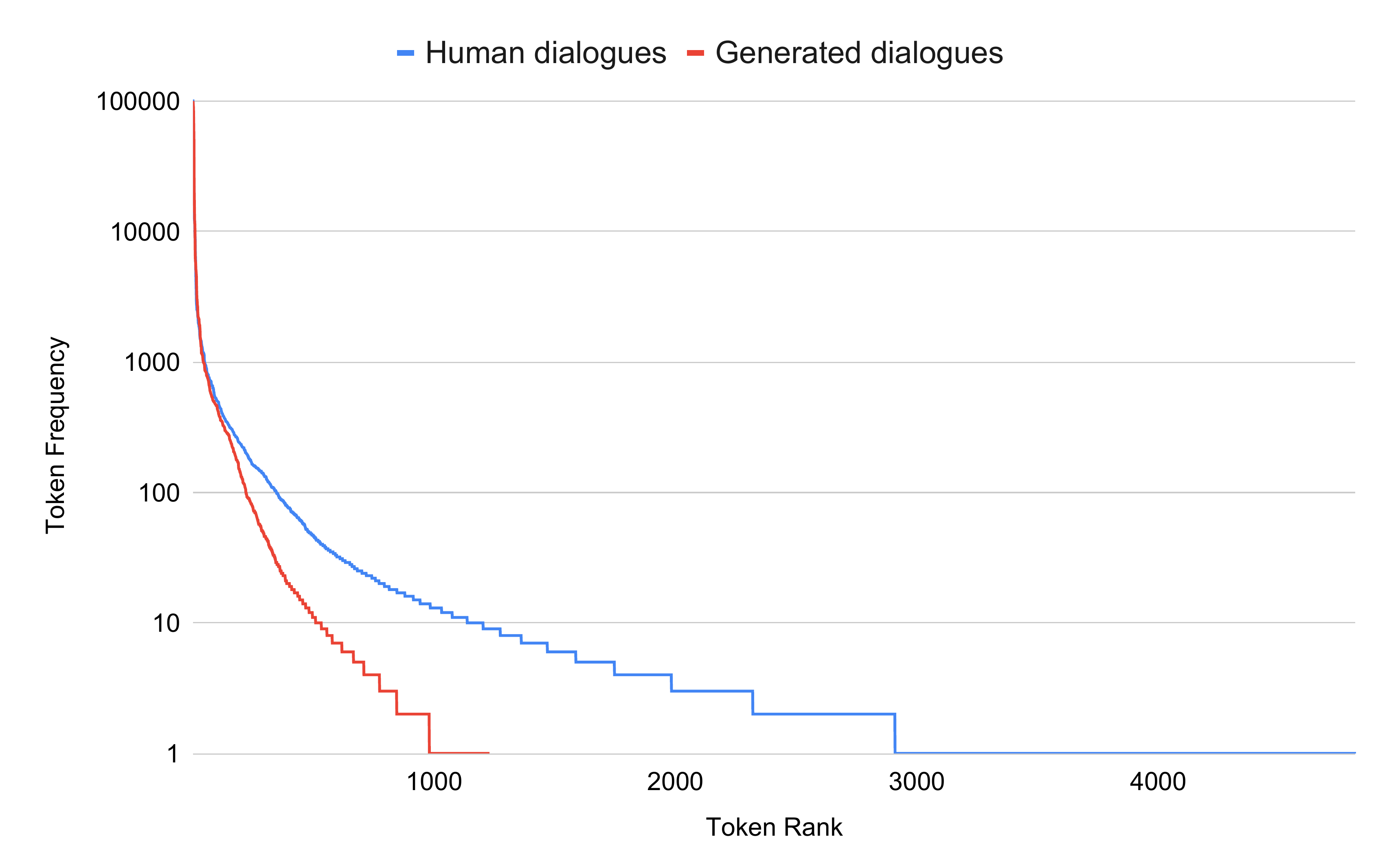}
	\caption{Token frequency plot (Zipf's Law curve) for human vs. generated dialogues}
	\label{fig:zipf}
	\vspace{0.35in}
\end{figure}

Figure \ref{fig:zipf} reports the token frequency curve for human dialogues and generated dialogues on the GuessWhat test set (Zipf's law). Human dialogues are clearly more rich and diverse compared to generated dialogues.

\begin{figure}[b!]
	\centering 
	\includegraphics[width=1\linewidth]{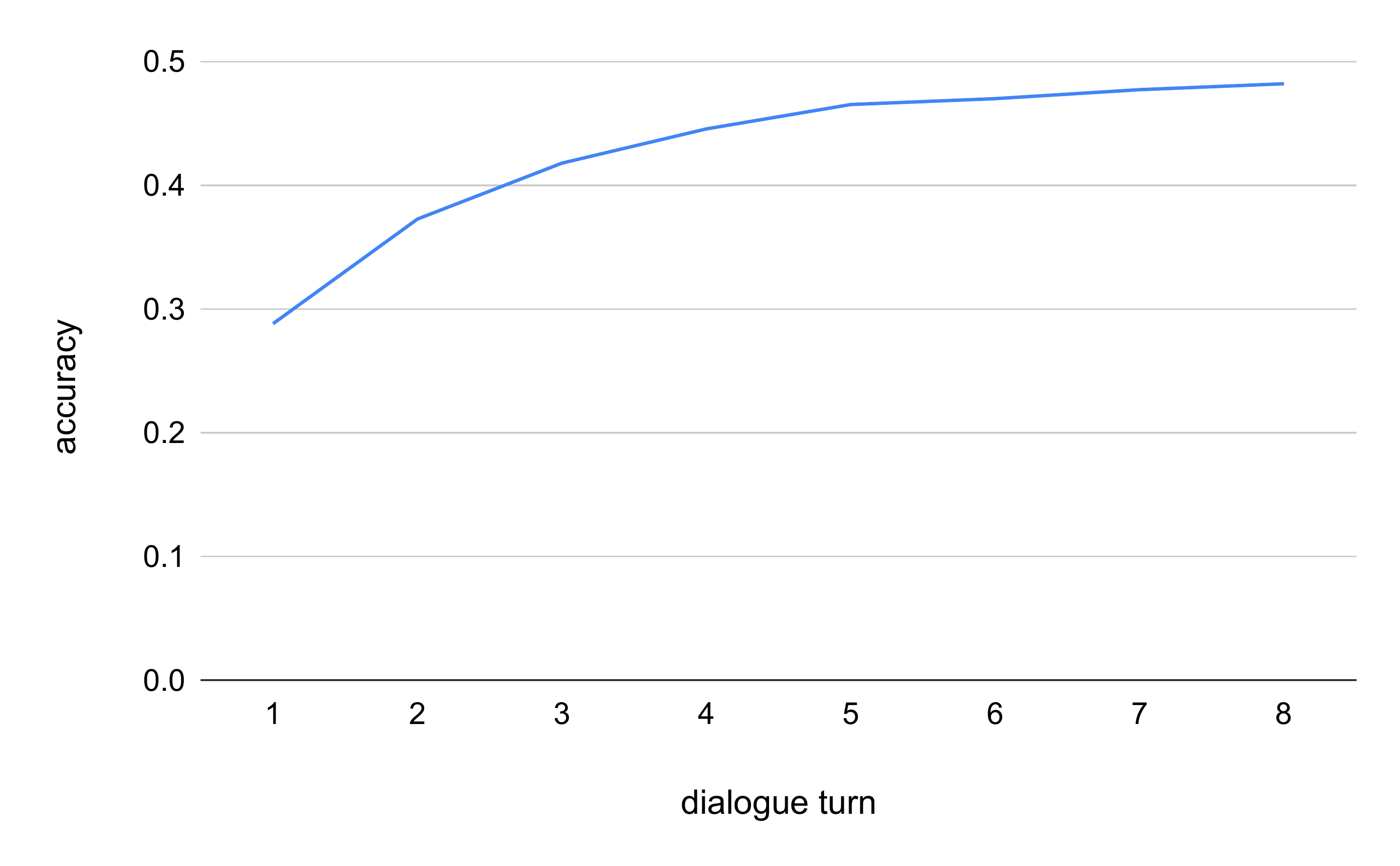}
	\caption{GDSE-SL per-turn accuracy on the GuessWhat game.}
	\label{fig:guesswhat_turn}
\end{figure}

\begin{figure}[t!]
	\centering 
	\includegraphics[width=1\linewidth]{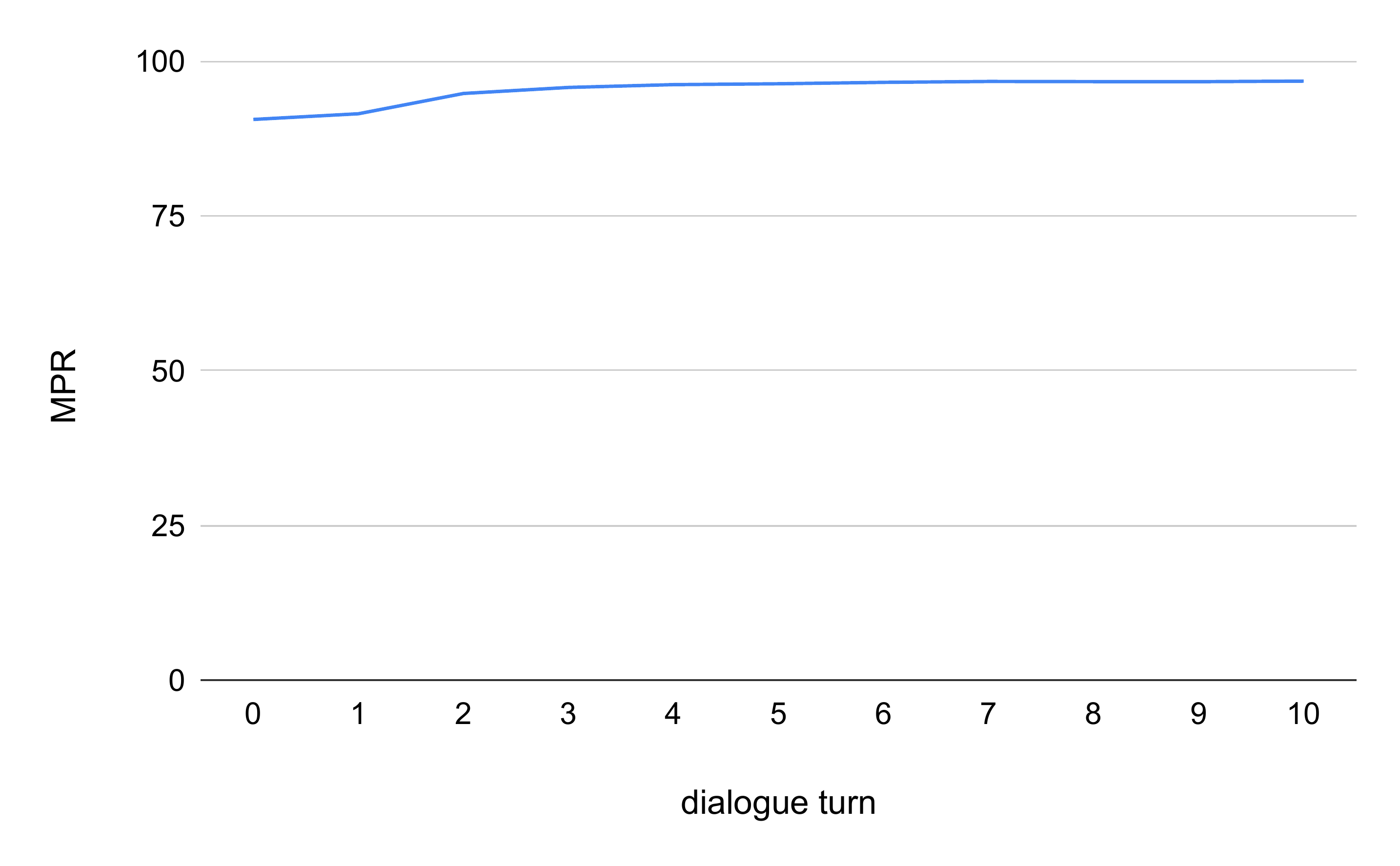}
	\caption{ReCap per-turn Mean Percentile Rank (MPR) on the GuessWhich game}
	\label{fig:guesswhich_turn}
	\vspace{8in}
\end{figure}

Figure \ref{fig:guesswhat_turn} and Figure \ref{fig:guesswhich_turn} show the  per-turn accuracy of GDSE-SL for GuessWhat and ReCap for GuessWhich, respectively. For GuessWhat, we report the simple task accuracy on the game, while for GuessWhich we use the Mean Percentile Rank; please refer to the main paper for additional details. For GuessWhat, the accuracy keeps increasing while new turns are given as input to the model. For GuessWhich, on the other hand, the Mean Percentile Rank (MPR) is pretty stable after very few turns and it is already high at turn 0, i.e. when only the caption is provided without any dialogue history.

\end{document}